\documentclass[runningheads]{llncs}
\usepackage[T1]{fontenc}
\usepackage{graphicx,verbatim}
\usepackage{subcaption}
\usepackage{xcolor} 
\usepackage{graphicx}
\usepackage{comment}
\usepackage{amsmath,amssymb} 
\usepackage{color}
\usepackage{paralist}
\usepackage{stackengine}
\usepackage{adjustbox}
\usepackage{caption} 
\usepackage{multirow}
\usepackage{diagbox}
\usepackage[colorlinks,linkcolor=blue]{hyperref}
\usepackage{bbm}
\usepackage{booktabs}
\usepackage{bm}
\usepackage{bbding}
\usepackage{colortbl} 
\usepackage{xcolor}   
\begin{document}
\title{Parameterized Diffusion Optimization enabled Autoregressive Ordinal Regression for Diabetic Retinopathy Grading}
%
\author{Qinkai Yu \inst{1} \and 
Wei Zhou \inst{2}\and 
Hantao Liu \inst{2}\and
Yanyu Xu \inst{3}\and
Meng Wang \inst{4}\and
Yitian Zhao \inst{5}\and 
Huazhu Fu \inst{6}\and
Xujiong Ye \inst{1}\and
Yalin Zheng \inst{7}\and Yanda Meng \inst{1 (}\Envelope\inst{)}}  
\authorrunning{Q.Yu et al.}

\institute{
\textsuperscript{$1$} Computer Science Department, University of Exeter, Exeter, UK\\
\textsuperscript{$2$} School of Computer Science and Informatics, University of Cardiff, Cardiff, UK\\
\textsuperscript{$3$} The Joint SDU-NTU Centre for Artificial Intelligence Research (C-FAIR), Shandong University, Jinan, China. \\ 
\textsuperscript{$4$} Centre for Innovation and Precision Eye Health \& Department of Ophthalmology, Yong Loo Lin School of Medicine, National University of Singapore, Singapore. \\
\textsuperscript{$5$} Ningbo Institute of Materials Technology and Engineering, Chinese Academy of Sciences, Ningbo, China \\
\textsuperscript{$6$} Institute of High Performance Computing, Agency for Science, Technology and Research, Singapore. \\
\textsuperscript{$7$} Eye and Vision Sciences Department, University of Liverpool, Liverpool, UK\\
\email{Y.M.Meng@exeter.ac.uk} \\
}




\maketitle              
\begin{abstract} 
As a long-term complication of diabetes, diabetic retinopathy (DR) progresses slowly, potentially taking years to threaten vision. An accurate and robust evaluation of its severity is vital to ensure prompt management and care. Ordinal regression leverages the underlying inherent order between categories to achieve superior performance beyond traditional classification. However, there exist challenges leading to lower DR classification performance: 1) The uneven distribution of DR severity levels, characterized by a long-tailed pattern, adds complexity to the grading process. 2)The ambiguity in defining category boundaries introduces additional challenges, making the classification process more complex and prone to inconsistencies. This work proposes a novel autoregressive ordinal regression method called AOR-DR to address the above challenges by leveraging the clinical knowledge of inherent ordinal information in DR grading dataset settings. Specifically, we decompose the DR grading task into a series of ordered steps by fusing the prediction of the previous steps with extracted image features as conditions for the current prediction step. Additionally, we exploit the diffusion process to facilitate conditional probability modeling, enabling the direct use of continuous global image features for autoregression without relearning contextual information from patch-level features. This ensures the effectiveness of the autoregressive process and leverages the capabilities of pre-trained large-scale foundation models. Extensive experiments were conducted on four large-scale publicly available color fundus datasets, demonstrating our model's effectiveness and superior performance over six recent state-of-the-art ordinal regression methods. The implementation code is available at \url{https://github.com/Qinkaiyu/AOR-DR}.

\keywords{Autoregressive ordinal regression  \and Diabetic retinopathy.}

\end{abstract}

\section{Introduction}


Ordinal regression not only learns categories but also explicitly models the ordinal relationships between them \cite{coral2020,shi2021deep} \cite{yu2024clip} \cite{li2022ordinalclip,liang2023crowdclip,li2021learning}  \cite{lee2022geometric}, which makes it ideal for classifying diabetic retinopathy (DR) \cite{kempen2004prevalence} into five ordinal groups: No DR, Mild, Moderate, Severe, and Proliferative. However, two major challenges arise: 1) Long-tailed data, where categories like Proliferative are underrepresented, weakening ordinal relationships; 2) Ambiguity of class boundaries, where differences in annotators' definitions lead to inconsistent boundaries. We propose that autoregressive methods can address these challenges by breaking the ordinal regression task into binary classification steps, where each prediction conditionally depends on the previous one. This approach inherently requires more inference steps to determine the outcome for the categories positioned further along the ordinal scale. In other words, these `further' categories (\textit{e.g.} severe, proliferative) will contribute more significantly to the overall loss optimization compared with other categories (\textit{e.g.} no DR, mild). This helps to compensate for the deficiencies caused by the under-representation of these `further' categories in the long-tailed DR distributions. Meanwhile, the autoregressive mechanism effectively leverages the transition characteristics between adjacent categories, enabling a gradual and stepwise prediction that enhances discrimination at the ambiguous boundaries between classes.

However, ordinal regression methods based on the autoregressive paradigm remain largely unexplored. This is mainly because the autoregressive mechanism typically uses patch-level features extracted by a backbone network, converting image patches into discrete tokens \cite{chen2020generative}. These features are then fed into a downstream decoder to capture contextual information across patches, a process commonly employed in object detection \cite{chen2021pix2seq} and image segmentation \cite{wang2024autoregressive}. However, backbone networks used in image classification tasks are often specifically designed to extract continuous global features \cite{dosovitskiy2020image}, which is contradictory to some extent. Ord2Seq \cite{wang2023ord2seq} follows the above approach by converting class labels into sequences with ordinal information and adopts a masked prediction strategy reminiscent of the BERT \cite{devlin2018bert} training paradigm, using a transformer to predict the sequence. Such an approach fails to fully leverage pre-trained models. It is worth noting that, the autoregressive paradigm has not been proposed for ordinal regression tasks. \cite{li2024autoregressive} breaks through aforementioned contradiction by enabling autoregressive training without vector quantization via their proposed diffusion process, enabling continuous global features to be utilized in downstream autoregressive tasks.

In this work, we propose a novel ordinal regression model called \textbf{AOR-DR} to address the challenges of long-tail distribution and ambiguity of class boundaries in diabetic retinopathy grading tasks. 
In particular, we introduce feature fusion to jointly condition image features with previous predictions. To this end, we adopt two widely used strategies: affine fusion and cross-attention fusion. Our proposed method represents the first attempt at an autoregressive paradigm for ordinal regression, exploiting diffusion-based optimization to model conditional probability distributions effectively without tokenization. 
Additionally, our work is highly compatible with mainstream backbone architectures, benefiting from large-scale pre-trained foundation models, such as RETFound \cite{zhou2023foundation}. 

The contributions of this work are as follows: 
\begin{itemize}
    \item We decompose the DR grading task into a series of ordered steps by fusing the prediction of the previous steps with extracted image features as conditions for the current prediction step.
    \item We exploit the diffusion process to facilitate conditional probability modeling, enabling the direct use of continuous global image features for autoregression without relearning contextual information from patch-level features. 
    \item Extensive experiments were conducted on four large-scale publicly available datasets, demonstrating our superior performance over six recently proposed state-of-the-art ordinal regression methods. We also observe that the classic autoregressive mechanism can bring a considerable performance boost solely on simple cross-entropy loss. However, its robustness may diminish across datasets with varying long-tailed distributions. In contrast, our proposed model consistently achieves optimal performance across all four dataset configurations.
\end{itemize}

\section{Methods}
\begin{figure*}[tb]
\centering
\includegraphics[width=0.95\textwidth]{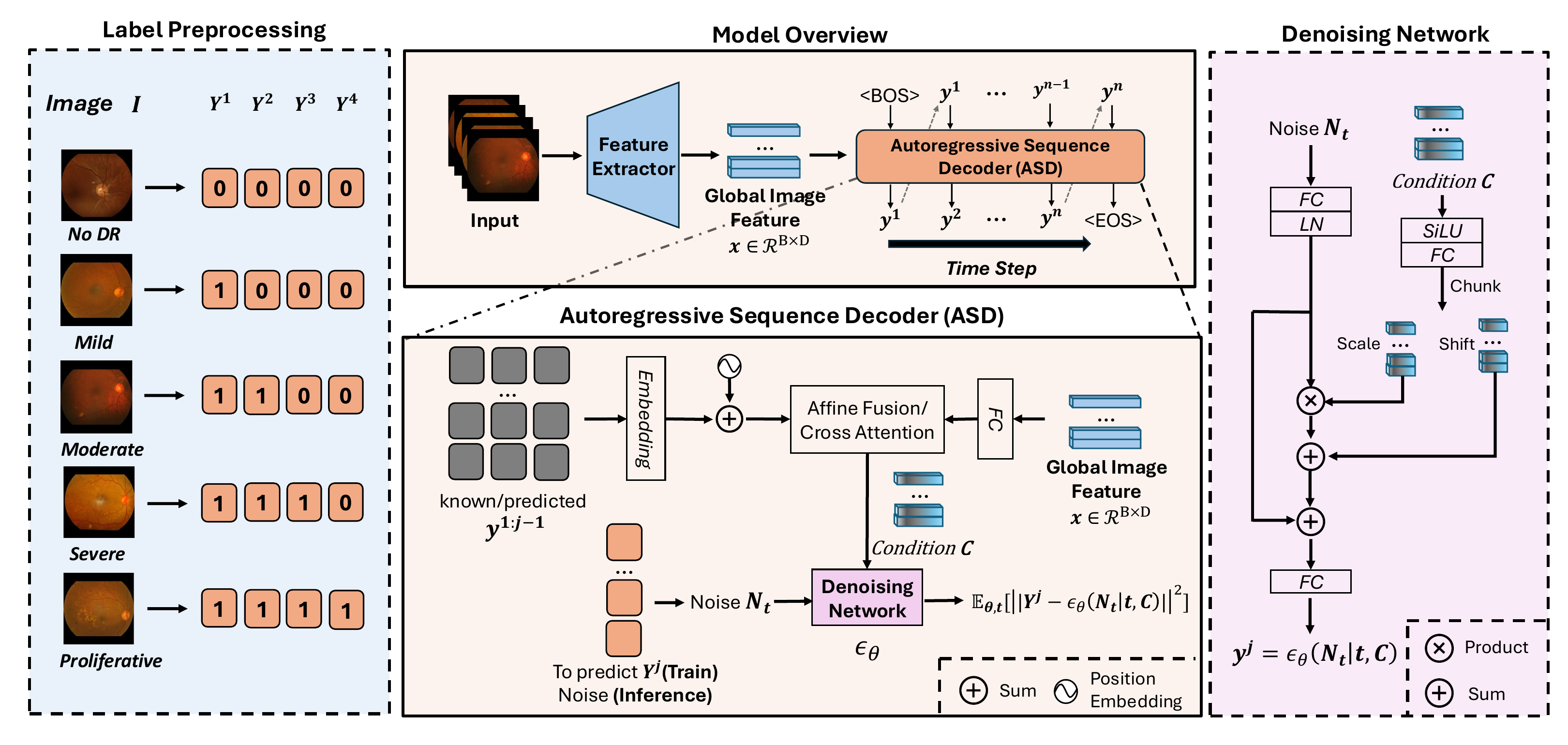}
\caption{Overview of the proposed AOR-DR framework. The Label Preprocessing part divides the DR grading task into
a series of ordered steps. Our model architecture consists of two main modules: feature extractor and autoregressive sequence decoder. The image feature extractor outputs global image features, denoted as $x \in \mathcal{R}^{B\times D}$. The autoregressive sequence decoder takes these global features $x$ along with the prediction from previous steps 
$y^{1:k-1}$ as input. The architecture of the denoising network is referenced by the final layer designed in DiT \cite{peebles2023scalable}. The condition $C$ is linearly transformed and decomposed into learnable scaling and shift, then merged with the noise $N_{t}$ through residual connections.}
\label{figure2}
\end{figure*}

An overview of the proposed AOR-DR framework is shown in \autoref{figure2}. The details of label preprocessing, model structure, and other proposed components are elaborated below.
\subsection{Preliminaries}
For a classification task involving $\textit{K}$ classes with images $\textit{I} \in \mathcal{R}^{C\times H \times W} $, our autoregressive framework decomposes the ordinary process into $\textit{K-1}$ binary classification tasks. It is represented as $f: I \rightarrow y^{1:k-1}$, where $y^{1:k-1}$ denotes $\{y^1 ,y^2 , ...,y^{K-1} \}$ and each $y^i \in \{0,1\}$ is a binary variable.  For example, the corresponding labels of `Severe' in DR grading (fourth class, the original label is 3) would be $\{ 1,1,1,0\}$. To align with the autoregressive paradigm, we introduce a start token <$BOS$> and an end token <$EOS$> at the beginning and end positions. The complete label in our training process becomes $Y = \{\text{<}BOS\text{>},Y^1 ,Y^2 , ...,Y^4, \text{<}EOS\text{>} \}$. Thus, for the given label $Y$, our proposed autoregressive model reformulates the classic ordinal regression as the following optimization objective: 
\begin{equation}
    \max p(y^{1},...,y^{4}) = \max \prod_{i=1}^4 p(y^i|y^1,...,y^{i-1}),
\label{eq1}
\end{equation}
where $p(\cdot)$ represents the probability that the model predicts the correct outcomes. 
\subsection{Model structure}
\noindent \textbf{Feature Extractor.} Our model is compatible with commonly used mainstream backbones. In this work, we adopt the RETFound \cite{zhou2023foundation}, a large foundation model pre-trained on 1.6 million color fundus images, as the feature extractor backbone. Specifically, we use vanilla ViT \cite{dosovitskiy2020image} (including ViT-B and ViT-L). The class token obtained by the ViT backbone is the global image feature $x$. The dimension $D$ of $x$ is 768 in ViT-B and 1024 in ViT-L. Even with such a powerful backbone network, our proposed autoregressive mechanism can still bring a large margin improvement over four large-scale datasets. 

\noindent \textbf{Autoregressive Sequence Decoder.} In this component, image features are fused with predictions from the previous steps, and diffusion-based optimization is adopted to model continuous fusion features as conditions to the current step prediction. Inspired by the approach proposed by \textit{Li et al.} \cite{li2024autoregressive}, we use an MLP with multiple residual structures as the denoising network in the diffusion process. 
Differently, we introduce feature fusion to jointly condition image features with previous predictions. To this end, we adopt two widely used strategies: affine fusion and cross-attention fusion. To ensure the effectiveness of the autoregressive process, we introduce the position embedding to represent the current step in the sequence, similar to those in vanilla ViT \cite{dosovitskiy2020image}. The prediction from the previous steps is passed through an embedding layer $h$, then combined with the corresponding position embedding $e$. We can define the embedded predictions $y^{j-1}_{embd}$ donate in $j-1$ step as: 
\begin{equation}
    y^{j-1}_{embd} = h(y^{1:j-1}) + e^{j-1}.
\end{equation}
Meanwhile, the global image features $x$, extracted from the backbone, are projected through a fully connected layer to match the same dimension as the embedded predictions from the previous step. The final fused features are fed into the denoising network as condition $C$ for the current step's prediction. The fusion process of affine fusion and cross-attention fusion is formulated as follows:
\begin{equation}
    C =\text{Condition}(x,y^{j-1}_{embd}) = FC(x)\cdot y^{j-1}_{embd} + FC(x),
\end{equation}
and \begin{equation}
    C = \text{Condition}(x,y^{j-1}_{embd}) = \text{Softmax}\frac{(W_{Q}\cdot FC(x))( W_{K}\cdot y^{j-1}_{embd} )}{\sqrt{d_{K}}} (   W_{V}\cdot y^{j-1}_{embd} ),
\end{equation}
where $W_Q$, $W_K$, and $W_V$ represent learnable weight matrices used to project the input image feature $x$ and the previous embedding $y^{j-1}_{embd}$ into query, key, and value spaces, respectively. The term $\sqrt{d_{K}}$ is a scaling factor. 
\subsection{Modeling Conditional Probabilities via Diffusion Process}
Denoising diffusion networks are considered capable of modeling arbitrary distributions \cite{ho2020denoising,song2020denoising}. A previous study \cite{yang2023diffmic} demonstrated the effectiveness of using diffusion models for modeling entropy, achieving remarkable performance in classification tasks. In our case, the diffusion models are employed to represent the distribution of each ordinal step prediction in the autoregressive process. Let $y^{j}$ be the prediction in step $j$. We reformulate the conditional probability optimization objective in \autoref{eq1} as: $\max$ $p(y^{j} |C)$ at step $j$, where $C$ combines previous predictions and image features. Here, we present the training and inference process in step $j$ as an example.

\noindent \textbf{Training.} The $y^j$ is fed into the denoising network after adding random Gaussian noise:     $N_{t} = \sqrt{\alpha_{t}}y^j+\sqrt{1-\alpha_{t}} \epsilon$, where $\epsilon$ is the noise sampled from $N(0,1)$; $\alpha$ defines the noise schedule and $t$ is the diffusion time step of the noise schedule. Modeling the conditional probability distribution $p(y^j|C)$ can be formulated with the following loss function:
\begin{equation}
    L(C,y^j)= \mathbb{E}_{\epsilon,t}\left [ \|Y^j -\epsilon_{\theta}(N_{t}|t,C) \|^2\right],
\label{diffusion loss}
\end{equation}
where $\epsilon_{\theta}$ represents the denoising network with parameter $\theta$. $\epsilon_{\theta}(N_{t}|t,C)$ means the network takes Noise $N_{t}$ as input and is conditional on both $t$ and $C$. 
In the end, \autoref{diffusion loss} is used as a diffusion-based parameterized optimization function during training. In this way, our proposed autoregressive mechanism directly leverages continuous global features to model conditional probabilities in continuous space, avoiding the limitations of traditional autoregressive paradigms that involve introducing additional transformer decoders to relearn contextual information from patch-level features, which may hinder the effective utilization of pre-trained models.
 \\
\noindent \textbf{Inference.} In the inference phase, predictions are obtained by sampling the distribution $p(y^{j}|C)$. To be specific, the sampling follows the reverse diffusion procedure, starting with $y^{j}_{t} \sim N(0,1)$, ending with $ y^{j}_{0} \sim p(y^j|C)$. Thus we can define $t-1$ step $y^{j}_{t-1}$ of the reverse diffusion process as:
\begin{equation}
    y^j_{t-1} = \frac{1}{\sqrt{\alpha_{t}}}(y^j_{t}-\frac{1-\alpha_{t}}{\sqrt{1-\alpha_{t}}}\epsilon_{\theta}(y^j_{t}|t,C) )+\sigma_{t}\delta,
\end{equation}
where $\delta$ is sampled from the Gaussian distribution $N(0,1)$, and $\sigma_{t}$ is the noise level at the diffusion time step $t$. Finally, we normalize the sampled $y^j_{0}$ to get the binary prediction $y^j$.
\section{Datasets and Implementation Details}\textbf{Evaluation Settings.} 
We conducted extensive experiments on four large-scale publicly available diabetic retinopathy grading datasets with various long-tail distribution characteristics, namely APTOS \cite{karthik2019aptos}, Messidor \cite{abramoff2016improved}, DDR \cite{li2019diagnostic}, DeepDR \cite{liu2022deepdrid}. The datasets were pre-split by the providers, and we used 20\% of the training set for validation. Performance was evaluated using Accuracy (ACC), Macro F1-Score (F1), Sensitivity (Sen), and Specificity (Spec), with the highest scores in \textbf{bold} and the second-highest in \underline{underline}. We compare the experimental results with the re-implemented state-of-the-art ordinal regression models with their publicly available code, including CORAL \cite{coral2020}, CORN \cite{shi2021deep}, PoEs \cite{li2021learning}, GoL \cite{lee2022geometric}, Ord2Seq \cite{wang2023ord2seq}, and CLIP-DR \cite{yu2024clip}.

\noindent\textbf{Implementation Details.} To ensure the 
fairness of the comparison, all compared models use the same backbone network ViT-B/16 \cite{dosovitskiy2020image} with the input images resized to 224 $\times$ 224. To mitigate the randomness introduced by diffusion predictions, we perform five times predictions for each sample and use their average as the final results. Image augmentations only include horizontal flipping. All models were trained with a batch size of 32 for the same number of epochs. For the diffusion model hyperparameter, the total number of diffusion steps was set to 1000 for training, while 100 steps were used for inference sampling to improve efficiency.
\begin{table*}[t!]
\caption{Comparison with recent state-of-the-art ordinal regression models.  * represents replacing the backbone as ViT-B/16. The best results are \textbf{bolded}, with \underline{underline} indicating the second highest.}
\renewcommand\arraystretch{1}
\centering 
\scalebox{0.70}{
\begin{tabular}{ccccccccccccccccc}
\toprule
\multirow{2}{*}{Model} & \multicolumn{4}{c}{\textbf{APTOS}} & \multicolumn{4}{c}{\textbf{Messidor}} & \multicolumn{4}{c}{\textbf{DDR}}&\multicolumn{4}{c}{\textbf{DEEPDR}}\\
\cmidrule(lr){2-5} \cmidrule(lr){6-9}\cmidrule(lr){10-13}\cmidrule(lr){14-17}
&\textit{ACC}& \textit{F1}&\textit{Sen}&\textit{Spec}  & \textit{ACC}  & \textit{F1}&\textit{Sen}&\textit{Spec}  & \textit{ACC} & \textit{F1}&\textit{Sen}&\textit{Spec} &\textit{ACC}&\textit{F1}&\textit{Sen}&\textit{Spec} \\
\midrule
ViT-B/16 & 69.9&38.3&36.4&68.6& \underline{63.9}&\underline{41.9}&43.2&72.7 &57.4& 26.0&22.3&63.6&62.5&39.1&39.8&71.7  \\
CORAL* \cite{coral2020}  & 66.4&30.4&27.3&67.8   &62.2 &18.7&29.0&72.3 &54.1& 26.6&24.7&67.2&57.7&37.3&36.9&71.2\\
CORN* \cite{shi2021deep} & 69.6& 31.4 &27.9&60.2&62.2  &18.6 &24.9 &72.3&54.7&26.3&23.9&66.5 &56.2&31.4&28.6&71.1\\
PoEs* \cite{li2021learning} & 67.2& 32.4 &30.4&68.2&44.7 &28.1  &31.8&71.2&51.3&28.0  &41.6&68.8&55.5&31.3&34.2&71.8\\
GoL* \cite{lee2022geometric}  & 74.2& 52.2 &\underline{60.7}&\underline{71.4}&60.8&33.8   &41.4&72.6&63.6&40.9  &46.1&\underline{69.3}&62.7&\underline{51.2}&56.7&72.8\\
Ord2Seq \cite{wang2023ord2seq}  & 70.2& 30.4 &27.7&64.8& \textbf{64.7} &34.1  &44.5&72.6&64.7&34.1&36.9&65.5 &61.2&43.2&52.4&71.3\\
CLIP-DR* \cite{yu2024clip}  & 71.8& 45.5 &44.5&71.1&63.3  &36.8  &45.8&72.7 &\underline{68.4}&\underline{45.6} &\underline{50.5}&69.2&64.5&50.6&\underline{57.1}&72.5\\
\midrule
\rowcolor{gray!20} Ours (Affine Fusion) & \underline{74.5}& \underline{52.5} &58.2&71.0&58.3 &37.6 &\textbf{52.2}&\underline{72.7}&\textbf{70.3}& \textbf{46.5} &\textbf{51.6}&68.3&\textbf{65.7}&50.5&47.6&\underline{72.5}\\
\rowcolor{gray!20} Ours (Cross Attention)&\textbf{77.1} &\textbf{58.6}&\textbf{62.6}&\textbf{71.9}&62.5&\textbf{45.1}&\underline{50.6}&\textbf{73.0}&65.0
 & 43.9 &49.5&\textbf{69.7}& 65.0&\textbf{51.9}&  \textbf{57.7}&\textbf{72.9}\\
\bottomrule[1pt]
\end{tabular}
\label{tab:1}
}
\end{table*}
\section{Results}
\label{result}
\begin{figure*}[t!]
\centering
\includegraphics[width=0.90\textwidth]{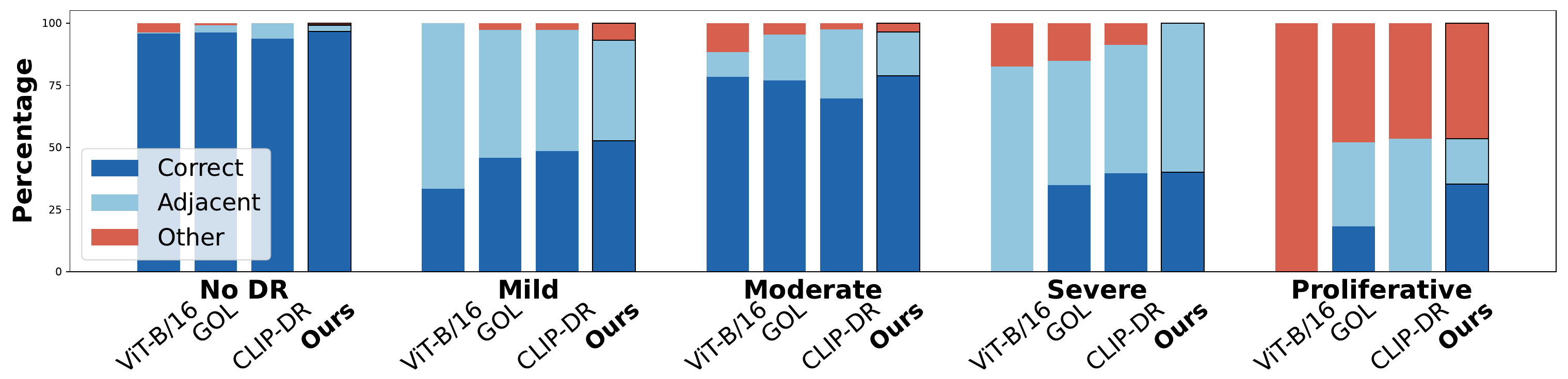}
\caption{The performance of the vanilla ViT-B/16 \cite{dosovitskiy2020image}, GOL \cite{lee2022geometric}, CLIP-DR \cite{yu2024clip}, and \textit{ours} on the APTOS \cite{karthik2019aptos} dataset is analyzed for each category, presenting the percentage of test dataset that classify \textbf{correctly}, misclassified into \textbf{adjacent} categories, and misclassified into \textbf{other} categories.}
\label{figure3}
\end{figure*}

\begin{figure}[t!]
    \centering\includegraphics[width=1\linewidth]{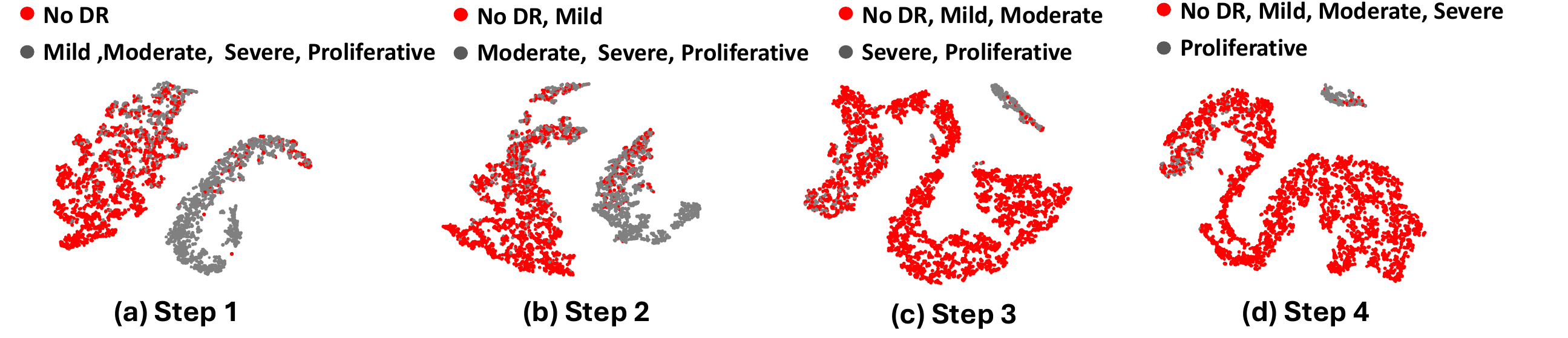}
    \caption{The t-SNE visualization illustrates our model's feature distribution before the final $FC$ layer on the DDR \cite{li2019diagnostic} test dataset. 
    }
    \label{fig:tsne}
\end{figure}

\textbf{Main Results.}
\autoref{tab:1} shows that our proposed model AOR-DR consistently outperforms other state-of-the-art methods across all evaluated datasets. Classic ordinal regression methods, such as k-rank classifiers-based models CORN \cite{shi2021deep} and CORAL \cite{coral2020} struggle with the long-tailed distribution of DR data, leading to suboptimal performance. Furthermore, other state-of-the-art methods cannot obtain consistent promising performance across the four datasets. For instance, the relative ordinal estimate-based algorithm, GoL \cite{lee2022geometric}, struggled to maintain good performance on the Messidor \cite{abramoff2016improved} test set due to the absence of certain categories. Similarly, PoEs \cite{li2021learning} performed poorly when confronted with the ambiguity of class boundaries (\textit{e.g.} `No DR' and `Mild' in APTOS \cite{karthik2019aptos} test dataset). Although Ord2Seq \cite{wang2023ord2seq} and CLIP-DR \cite{yu2024clip} demonstrated relatively robust grading performance, they fall short of achieving optimal performance compared to our proposed model. Specifically, our model outperforms Ord2Seq \cite{wang2023ord2seq} 42 \% on \textit{F1} and 51 \%  on \textit{Sen} in average and outperforms CLIP-DR \cite{yu2024clip} 13 \% on \textit{F1} and 13 \%  on \textit{Sen} in average. To further demonstrate our model's superior ability to handle long-tail distributions and class ambiguities in DR grading, we compared its accuracy in predicting correct, adjacent, and other classes against vanilla ViR-B/16 \cite{dosovitskiy2020image}, GoL \cite{lee2022geometric} and CLIP-DR \cite{yu2024clip} on the APTOS \cite{karthik2019aptos} dataset. As shown in \autoref{figure3}, our model demonstrates superior performance across all categories, particularly notable improvements in the underrepresented classes, \textit{e.g.} severe and proliferative. Additionally, \autoref{fig:tsne} shows the t-SNE visualization results of our model's representation before the final $FC$ layer on the DDR \cite{li2019diagnostic} test dataset. Clear decision boundaries separating neighboring classes are observed at each step, where the expanding red area represents the set of samples whose ordinal grade has been conclusively determined, validating the effectiveness of the proposed autoregressive model. A potential concern with this autoregressive setup is the possibility of generating invalid ordinal sequences (e.g., [0, 1, 0, 1]). However, since the training dataset contains no such invalid constructs, the model implicitly learns to avoid them. Empirically, we did not observe any invalid sequences in the test set after the model was fully trained.

\noindent \textbf{Ablation study on different backbones.}
Our proposed method can be integrated into widely used backbone architectures. We conduct ablation study experiments and evaluate its performance on the ViT \cite{dosovitskiy2020image} series, including ViT-B/16, and RETFound (ViT-L/16)\cite{zhou2023foundation}, shown in \autoref{tab:ablation}. Notably, for RETFound \cite{zhou2023foundation}, the backbone is frozen during training to demonstrate that our method can leverage the performance of pre-trained models, while for ViT-B, full parameters fine-tuning is performed. In all test dataset scenarios, \textit{Ours} demonstrates superior adaptability and performance, consistently achieving promising performance under all four metrics. Relatively modest performance improvement is observed for the comparison with RETFound \cite{zhou2023foundation} due to the frozen backbone weights. 
\noindent \textbf{Ablation study on different optimization functions.} 
To verify the effectiveness of the diffusion-based optimization function, we remain the model structure unchanged and replace the proposed parameterized diffusion loss with other commonly used loss functions, such as cross-entropy loss. The fusion strategy was selected based on the best-performing approach for each test dataset in \autoref{tab:1}. Results in \autoref{tab:ablation} indicate that on the APTOS \cite{karthik2019aptos} and DDR \cite{li2019diagnostic} datasets, the cross-entropy based autoregressive method outperforms ViT-B/16. However, on the Messidor \cite{abramoff2016improved} and DEEPDR \cite{liu2022deepdrid} datasets, which have severe long-tailed distributions with highly underrepresented classes, cross-entropy loss underperforms the baseline. This is likely because the autoregressive approach becomes trapped in repetitive patterns, hindering the learning of rare class features. Additionally, when training with the Retfound backbone frozen, the cross-entropy optimization fails to effectively leverage pretrained weights. In contrast, our proposed paradigm benefits from a diffusion-based optimization procedure that operates in continuous space, allowing for more effective conditional probability modeling of the global continuous features and thereby achieving superior performance.
\begin{table*}[t]
\caption{DR grading performance with different backbone network structures and different optimization functions. $\dagger$ represents freezing the backbone during training. The best results are \textbf{bolded}, with \underline{underline} indicating the second highest.}
\renewcommand\arraystretch{1}
\centering 
\scalebox{0.67}{
\begin{tabular}{ccccccccccccccccc}
\toprule
\multirow{2}{*}{Method} & \multicolumn{4}{c}{\textbf{APTOS}} & \multicolumn{4}{c}{\textbf{Messidor}} & \multicolumn{4}{c}{\textbf{DDR}}&\multicolumn{4}{c}{\textbf{DEEPDR}}\\
\cmidrule(lr){2-5} \cmidrule(lr){6-9}\cmidrule(lr){10-13}\cmidrule(lr){14-17}
&ACC& F1&Sen&Spec  & ACC  & F1&Sen&Spec  & ACC & F1&Sen&Spec &ACC&F1&Sen&Spec \\
\bottomrule[1pt]
ViT-B/16  & 69.9&38.3&36.4&68.6 &\underline{63.9}&\underline{41.9}&\underline{43.2}&\underline{72.7} &57.4& 26.0&22.3&63.6&\underline{62.5}&\underline{39.1}&\underline{39.8}&\underline{71.7}  \\
AR+ce loss (\textit{w/}ViT-B/16)& \underline{73.4}&\underline{54.4}&\underline{56.7}&\underline{71.1} &59.1&35.3&55.2&72.9 &\underline{67.4}&\underline{43.8}&\underline{48.3}&\textbf{69.8}&50.7&30.2&33.2&70.4  \\
\rowcolor{gray!20} Ours (\textit{w/}ViT-B/16)& \textbf{77.1} &\textbf{58.6}&\textbf{62.6}&\textbf{71.9}&\textbf{62.5}& \textbf{45.1}& \textbf{50.6}&\textbf {73.0}&\textbf{70.3}& \textbf{46.5}& \textbf{51.6} &\underline{68.3}&\textbf{65.0}&\textbf{51.6}&\textbf{57.7}&\textbf{72.9}\\
\bottomrule[1pt]
$\text{RETFound}^{\dagger}$ \cite{zhou2023foundation} &\textbf{82.1}&\underline{65.5}&\textbf{69.2}&\underline{72.6}& \underline{64.2}&\underline{45.4}&\underline{57.0}&\underline{72.9} &\underline{71.7}&\textbf{57.2}&\underline{50.3}&67.3&\underline{75.7}&\underline{57.3}&\underline{59.8}&\underline{73.9}  \\
AR+ce loss (\textit{w/}$\text{RETFound}^{\dagger}$) & {63.0}&{40.3}&{45.8}&70.9 &58.3&37.4&38.0&72.8&64.6&46.5&50.2&\textbf{69.8}&49.75&34.8&38.9&71.4 \\
\rowcolor{gray!20} Ours (\textit{w/}$\text{RETFound}^{\dagger}$) &\underline{80.3}&\textbf{65.7} &\underline{66.5}&\textbf{72.8}&\textbf{64.7} &\textbf{54.0} &\textbf{62.4}&\textbf{73.9}&\textbf{72.8}&55.9 &\textbf{55.3}&\underline{69.1}&\textbf{75.9}&\textbf{58.1}&\textbf{59.9}&\textbf{73.9}\\
\bottomrule[1pt]
\end{tabular}
\label{tab:ablation}
}
\end{table*}
\section{Conclusion}
We propose a novel ordinal regression model to address the challenges of long-tail distribution and inter-class ambiguity in the diabetic retinopathy grading task. Our work is the first attempt to introduce an autoregressive mechanism to ordinal regression, combined with a parameterized diffusion optimization process, enabling the direct utilization of continuous image features without traditional tokenization. Extensive experiments were conducted on four large-scale publicly available datasets, demonstrating the effectiveness and superior performance of our proposed approach, compared with six recent state-of-the-art ordinal regression methods.
\begin{credits}
\subsubsection{Acknowledgments}
This work is supported by Y. Meng's The Royal Society Fund (IEC\textbackslash
NSFC\textbackslash
242172), United Kingdom, and by H. Fu’s Agency for Science, Technology and Research (A*STAR) Central Research Fund (“Robust and Trustworthy AI system for Multi-modality Healthcare”). Qinkai Yu thanks the PhD studentship funded by Liverpool Centre for Cardiovascular Science and University of Exeter.
\end{credits}

\begin{credits}
\subsubsection{\discintname}
The authors have no competing interests to declare that
are relevant to the content of this article.
\end{credits}

%
%
\bibliographystyle{splncs04}
\bibliography{paper-0874}
%




\end{document}